\def\year{2021}\relax
\documentclass[letterpaper]{article} 
\usepackage{aaai21}  
\usepackage{times}  
\usepackage{helvet} 
\usepackage{courier}  
\usepackage[hyphens]{url}  
\usepackage{graphicx} 
\usepackage{natbib}  
\usepackage{caption} 
\usepackage[utf8]{inputenc} 
\usepackage{booktabs}       
\usepackage{amsfonts}       
\usepackage{nicefrac}       
\usepackage{microtype}      

\usepackage{adjustbox}
\usepackage{bm}
\usepackage{amsmath}
\usepackage{amssymb}
\newcommand{\ie}{\textit{i}.\textit{e}.}
\newcommand{\eg}{\textit{e}.\textit{g}.}

\usepackage[parfill]{parskip}

\usepackage{mathtools}
\DeclarePairedDelimiterX{\infdivx}[2]{(}{)}{%
  #1\;\delimsize\|\;#2%
}

\usepackage{pifont}
\usepackage{arydshln}
\newcommand{\cmark}{\ding{51}}
\newcommand{\xmark}{\ding{55}}

\usepackage{multirow}
\usepackage{amsmath,amssymb}
\usepackage{bbm}
\usepackage{dsfont}
\usepackage{amsthm}

\usepackage{xcolor}
\definecolor{red}{rgb}{1,0,0}
\newcommand{\red}[1]{\textcolor{red}{#1}}
\definecolor{blue}{rgb}{0,0,1}
\newcommand{\blue}[1]{\textcolor{blue}{#1}}
\definecolor{darkgreen}{rgb}{0,0.6,0}
\newcommand{\darkgreen}[1]{\textcolor{darkgreen}{#1}}
\definecolor{lightgray}{rgb}{.83,.83,.83}
\definecolor{gray}{rgb}{.75,.75,.75}

\title{Hypothesis Disparity Regularized Mutual Information Maximization}

\author {
        Qicheng Lao, \textsuperscript{\rm 1 \rm 2 \rm 3}
        Xiang Jiang, \textsuperscript{\rm 3}
        Mohammad Havaei \textsuperscript{\rm 3} \\
}
\affiliations {
    \textsuperscript{\rm 1} West China Biomedical Big Data Center, West China Hospital of Sichuan University, Chengdu, China \\
    \textsuperscript{\rm 2} MILA, Universit\'e de Montr\'eal, Montreal, Canada \\
    \textsuperscript{\rm 3} Imagia,\thanks{A US provisional patent application has been filed for protecting at least one part of the innovation disclosed in this article. This work was done at Imagia, funded through MEDTEQ grant 10-30 IA Multicentriq.} Montreal, Canada \\
    qicheng.lao@gmail.com, xiang.jiang@imagia.com, mohammad@imagia.com
}

\begin{document}

\maketitle

\begin{abstract}
We propose a hypothesis disparity regularized mutual information maximization~(HDMI) approach to tackle unsupervised hypothesis transfer---as an effort towards unifying hypothesis transfer learning (HTL) and unsupervised domain adaptation (UDA)---where the knowledge from a source domain is transferred solely through hypotheses and adapted to the target domain in an unsupervised manner. In contrast to the prevalent HTL and UDA approaches that typically use a single hypothesis, HDMI employs multiple hypotheses to leverage the underlying distributions of the source and target hypotheses. To better utilize the crucial relationship among different hypotheses---as opposed to unconstrained optimization of each hypothesis independently---while adapting to the unlabeled target domain through mutual information maximization, HDMI incorporates a hypothesis disparity regularization that coordinates the target hypotheses jointly learn better target representations while preserving more transferable source knowledge with better-calibrated prediction uncertainty. HDMI achieves state-of-the-art adaptation performance on benchmark datasets for UDA in the context of HTL, without the need to access the source data during the adaptation. 
\end{abstract}

\section{Introduction}
Mutual information (MI) maximization has been shown as a promising approach in unsupervised learning, as manifested in discriminative clustering~\cite{bridle1992unsupervised,krause2010discriminative} and unsupervised representation learning~\cite{hu2017learning,hjelm2018learning,tian2019contrastive,oord2018representation}. Recently, it has also been applied to unsupervised domain adaptation (UDA), achieving new state-of-the-art performance even in the more restricted context of hypothesis transfer learning (HTL \footnote{HTL was first introduced in~\cite{kuzborskij2013stability}.}) where the transfer of the knowledge from a source domain to a target domain is achieved solely through hypotheses~\cite{liang2020we}. Hypothesis transfer has the notable privacy-preserving property that respects the privacy of the source domain by eliminating the need to access the source data while transferring knowledge to the target domain, and is favored by both theoretical analysis~\cite{kuzborskij2013stability,ben2013domain,perrot2015theoretical,kuzborskij2017fast} 
and many empirical applications~\cite{fei2006one,yang2007cross,orabona2009model,jie2011multiclass,tommasi2013learning,du2017hypothesis,fernandes2019hypothesis}. However, HTL has been mostly explored in the supervised learning setting where the target labels are available except the work in~\cite{liang2020we}.

\begin{figure*}[!t]
    \centering
	\includegraphics[width=0.93\textwidth]{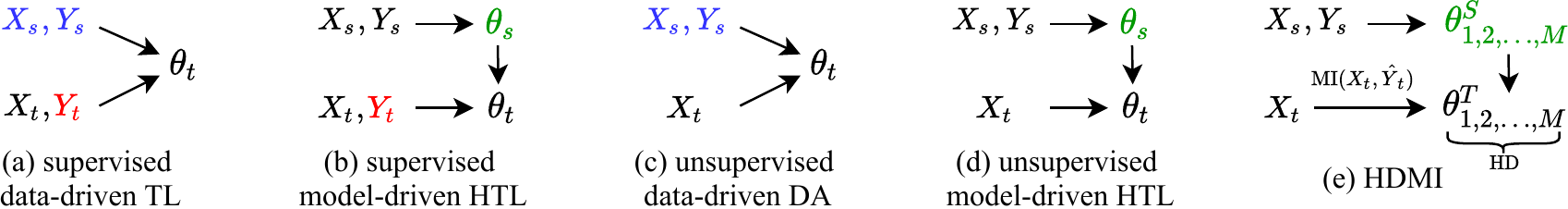}
	\caption{Graphical models for transfer learning. (a)-(d) Settings of transfer learning depending on the transfer medium and the availability of target labels; (e) Our proposed HDMI for unsupervised hypothesis transfer learning. The colors indicate direct access to the \blue{source data}, \red{target labels} and \darkgreen{source hypotheses} during target adaptation. }
	\label{fig:settings}
\end{figure*}

In this paper, we further tackle the problem of unsupervised hypothesis transfer (Figure~\ref{fig:settings} (d)) under the umbrella of MI maximization, as an effort to bridge the gap between HTL (Figure~\ref{fig:settings} (b)) and UDA (Figure~\ref{fig:settings} (c)). In contrast to the prevalent approaches for HTL and UDA that tend to use a single hypothesis failing to uncover different modes of the model distribution, we propose to transfer knowledge from a set of \textit{source hypotheses} learned from the source domain to a corresponding set of \textit{target hypotheses} by means of MI maximization on the unlabeled target data. The employment of multiple hypotheses is especially relevant to domain adaptation with out-of-distribution examples, and has a pronounced impact on the uncertainty calibration as well as the final adaptation or transfer performance, as has been demonstrated in previous work such as deep ensembles~\cite{lakshminarayanan2017simple}.


Furthermore, we highlight the crucial limitation of multiple independent MI maximization where different target hypotheses can be optimized in an unconstrained manner due to the lack of supervision, resulting in undesirable disagreements on the target label predictions as well as instability during the optimization process. 
To overcome such limitations and better take advantage of the relationship among different hypotheses, we propose a hypothesis disparity (HD) regularization to align the target hypotheses in a way such that the uncertainty manifested through different source hypotheses is taken into account while the undesirable disagreements are marginalized out. The HD regularization also shares the similar idea in unsupervised discriminative clustering with regularized information maximization, where a complexity penalty term is shown to be indispensable~\cite{krause2010discriminative}. We term the proposed multiple hypotheses MI maximization with HD regularization as Hypothesis Disparity regularized Mutual Information maximization (HDMI), illustrated in Figure~\ref{fig:settings}~(e). 

Finally, we evaluate the proposed HDMI approach on three benchmark datasets for domain adaptation in the context of unsupervised hypothesis transfer. We show that (i) the proposed HD regularization is effective in minimizing the undesirable disagreements among different target hypotheses and stabilizing the MI maximization process; (ii) Compared to direct MI maximization with single hypothesis or multiple hypotheses, the HD regularization facilitates the positive transfer of multiple modes from source hypotheses, and as a result, the target hypotheses obtained by HDMI preserve more transferable knowledge from each source hypothesis; (iii) HDMI uses well-calibrated predictive uncertainty to achieve effective MI maximization; and (iv) HDMI works through learning better representations shared by different target hypotheses. Overall, HDMI achieves new state-of-the-art performance in unsupervised hypothesis transfer learning.

\section{Related Work}
\paragraph{Relation to hypothesis transfer learning} Approaches for transfer learning can be categorized into data-driven approaches (\eg, instance weighting, feature transformation) and model-driven approaches~\cite{pan2009survey,zhuang2019comprehensive,fernandes2019hypothesis}. The differences between the two categories are illustrated in Figure~\ref{fig:settings} (a) and (b) in a supervised transfer learning setting. In this work, we focus on the model-driven category, which is also referred to as HTL in the literature~\cite{kuzborskij2013stability}. HTL was first introduced in~\cite{kuzborskij2013stability} with theoretical analysis on a regularized least squares problem, and later on extended to a general regularized empirical risk minimization problem~\cite{kuzborskij2017fast}. Approaches for HTL have also been proposed based on transformation functions~\cite{du2017hypothesis} and model structure similarity~\cite{fernandes2019hypothesis}. However, all the previous HTL approaches assume access to the labeled data in the target domain, \ie, supervised HTL, whereas we explore the possibility of unsupervised HTL in this work.

\paragraph{Relation to unsupervised domain adaptation} Unsupervised domain adaptation, also considered as a form of transfer learning (transductive transfer learning~\cite{pan2009survey}), aims to adapt a target domain to a source domain without requiring target labels, and has also been extensively studied in the past few years~\cite{ganin2015unsupervised,long2015learning,tzeng2017adversarial,hoffman2017cycada,saito2018maximum,pmlr-v97-zhang19i}. While most work assumes simultaneous access to both source and target data during adaptation, a few argue against the reality and necessity of such assumption~\cite{chidlovskii2016domain,liang2019distant,liang2020we}. We follow the setting of~\cite{liang2020we}, where the source data is inaccessible during target adaptation, and the source knowledge is transferred solely through hypotheses.

\paragraph{Relation to mutual information maximization}
The mutual information maximization can be achieved between input and output~\cite{bridle1992unsupervised,krause2010discriminative}, between input and intermediate representation or context~\cite{hu2017learning,hjelm2018learning,oord2018representation}, or between representations from different views~\cite{tian2019contrastive,bachman2019learning}. In this work, we maximize the MI between the input data of the target domain and the corresponding pseudo-labels predicted by the target hypothesis. In addition, we extend MI maximization to multiple hypotheses and also introduce a regularization term for MI maximization with multiple hypotheses. Unlike the regularized information maximization~\cite{krause2010discriminative} with the penalty term on the complexity of a single hypothesis, we emphasize on the disparity among multiple hypotheses; Compared with previous work in HTL that places the regularization directly between the source and target hypotheses~\cite{kuzborskij2013stability,fernandes2019hypothesis}, our proposed HD regularization for MI maximization is an indirect and relaxed form of regularization that is only among the target hypotheses. 

MI maximization has also been demonstrated to have better performance in discriminative clustering~\cite{krause2010discriminative}, compared with conditional entropy minimization~\cite{grandvalet2005semi} that was proposed for semi-supervised learning. In domain adaptation, approaches have been proposed based on MI maximization~\cite{liang2020we} and conditional entropy minimization, \eg, integrating with a minmax game~\cite{saito2019semi}, virtual adversarial training \footnote{VAT was originally proposed in~\cite{miayto2017virtual} for semi-supervised
text classification.}~\cite{shu2018dirt,kumar2018co}, or correlation alignment~\cite{morerio2017minimal}. HDMI is closely related to~\cite{liang2020we} in the sense that both are MI-based; however, instead of using a pseudo-label based self-training strategy to overcome the limitations of MI maximization in UDA, we propose to directly improve on MI maximization itself.

\paragraph{Relation to auxiliary classifiers and ensemble methods} 
The multiple hypotheses used in HDMI also have a connection to auxiliary classifiers used in multi-task learning~\cite{ruder2017overview}, domain adaptation from multiple sources~\cite{mansour2009domain,duan2009domain,peng2019moment} and hypothesis transfer from auxiliary hypotheses~\cite{yang2007cross,tommasi2013learning,kuzborskij2017fast}. While the auxiliary classifiers aim to leverage the knowledge learned from multiple different tasks or sources, our multiple hypotheses in this work focus on covering different modes of the underlying hypothesis distribution learned from a single source domain for a single task. In addition, HDMI also shares some appealing properties with ensemble methods, such as deep ensemble that improves uncertainty calibration~\cite{lakshminarayanan2017simple,fort2019deep}, and ensemble adversarial training for robustness~\cite{tramer2017ensemble}. However, HDMI differs from ensemble methods by exploiting a hypothesis disparity regularization during the unsupervised optimization process, and we show later in our experiments that two hypotheses suffice HDMI to benefit from this regularization.

\section{Approach}
\subsection{Problem formalization}
Let $\mathcal{X}$, $\mathcal{Y}$, $\mathcal{H}$ be the input space, the output space, and the hypothesis space, respectively. We denote the source domain data as $\mathcal{D}_s=\{(x_i^S, y_i^S)\}_{i=1}^{N_s}$, where $x_i^S \in \mathcal{X}^S$ and $y_i^S \in \mathcal{Y}^S$. Similarly, the unlabeled target data is denoted as $\mathcal{D}_t=\{(x_i^T)\}_{i=1}^{N_t}$, where $x_i^T \in \mathcal{X}^T$ and $\mathcal{X}^T \neq \mathcal{X}^S$.
In unsupervised HTL, the assumption is that the learning task $\mathcal{T}$ remains the same between the source and target domain where $\mathcal{Y}^S = \mathcal{Y}^T$ and $P_S(Y|X) = P_T(Y|X)$, which is also a typical assumption in closed-set UDA~\cite{panareda2017open}. Suppose a source hypothesis $h_s: \mathcal{X}^S \to \mathcal{Y}^S$ and a target hypothesis $h_t: \mathcal{X}^T \to \mathcal{Y}^T$, we have the following posterior predictive distribution from a Bayesian perspective:
\begin{equation} \label{eq_predictive}
\resizebox{\columnwidth}{!}{$
    p(Y_t^*|\mathcal{D}_t, \mathcal{D}_s) = \int_{h_t} p(Y_t^*|\mathcal{D}_t, h_t) \int_{h_s} p(h_t|\mathcal{D}_t, h_s) p(h_s|\mathcal{D}_s) dh_s dh_t,
$}
\end{equation}
with the goal to predict the unobserved target labels $Y_t^*$ by marginalizing over the posterior probabilities of both source and target hypotheses. Note that the target posterior $p(h_t|\mathcal{D}_t, h_s)$ is consistent with Figure \ref{fig:settings}~(d) where $h_t$ is conditioned on $h_s$ and $\mathcal{D}_t$, without direct access to the source data $\mathcal{D}_s$.
In contrast to the prevalent UDA approaches that assume $h_t = h_s$ and HTL approaches that typically use a single source and target hypothesis \footnote{Cases with multiple source domains are not included here.}, we propose to employ multiple hypotheses to better utilize the underlying distributions of source hypothesis $h_s$ and target hypothesis $h_t$.

Below we describe the proposed approach that leverages the idea of multiple hypotheses, \ie, given multiple source hypotheses trained on the source domain, we extract target knowledge from the unlabeled target data via mutual information maximization, with the constraint of minimized target hypothesis disparity, thus resulting in HDMI.

\subsection{Learning multiple source hypotheses} \label{sec_source}
As a first step, we learn a set of source hypotheses $\{h_i^S\in \mathcal{H}^S: \mathcal{X}^S \to \mathcal{Y}^S \}_{i=1}^{M}$ on the source data~$\mathcal{D}_s$.
We consider a set of source hypotheses $\{h_i^S: h_i^S = f_i^S \circ \psi^S \}_{i=1}^{M}$ that use a shared feature extractor $\psi^S$ but $M$ independent classifiers $\{f_i^S\}_{i=1}^{M}$ trained from different random initialization. Learning multiple source hypotheses is similar to \textit{deep ensemble}~\cite{fort2019deep} with the notable ability to learn diverse functions from different modes, whereas single hypothesis learning with maximum a posteriori only uncovers a single mode of the posterior $p(h_s|\mathcal{D}_s)$. In addition, compared with approximate Bayesian approaches like Monte Carlo dropout (MC-dropout)~\cite{gal2016dropout}, deep ensemble with random initialization is shown to produce well-calibrated uncertainty estimations that are more robust to domain shift~\cite{lakshminarayanan2017simple}.
This is especially relevant to HTL and has a pronounced impact on the transfer performance.


For ease of exposition, given an input $x$, we denote $h(x) = p(y|x; h)$ as the output label probability distribution predicted by a hypothesis $h$, where $y \in \{1,\dots K\}$ and $K$ is the number of classes. The multiple source hypotheses are learned jointly by minimizing the following objective function:
\begin{equation} \label{eq_source_loss}
\begin{split}
    & \mathcal{L}_{source} = \mathbb{E}_{h \in \mathcal{H}^S, (x, y) \in \mathcal{X}^S \times \mathcal{Y}^S} \left [ \ell_{\text{CE}} (h(x), y)  \right ],
\end{split}
\end{equation}
where $\ell_{\text{CE}}$ denotes the cross entropy loss function.
In practice, we use the average of $M$ hypotheses to approximate the expectation in Eq.~\ref{eq_source_loss}.

\subsection{Learning target hypotheses via mutual information maximization} \label{sec_target}
Given the unlabeled target data $\mathcal{D}_t=\{(x_i^T)\}_{i=1}^{N_t}$ and a set of previously learned source hypotheses $\{h_i^S\}_{i=1}^{M}$, we aim to adapt the source hypotheses into a set of corresponding target hypotheses $\{h_i^T\in \mathcal{H}^T\}_{i=1}^{M}$ by maximizing the mutual information between the empirical target input distribution and the predicted target label distribution induced by the target hypotheses.
Let $X^T$ denote the random variable for the target input, and $\hat{Y}^T$ denote the random variable of the model prediction inferred from hypothesis $h$ with the empirical label distribution $p(\hat{y}^T;h)=\frac{1}{N}\sum_{i} p(y|x_i^T; h)$. The MI between the target input $X^T$ and the output $\hat{Y}^T$ 
can be written as:
\begin{equation} \label{eq_MI}
\begin{aligned}
    I(X^T; \hat{Y}^T) & = H(\hat{Y}^T) - H(\hat{Y}^T | X^T). \\
\end{aligned}
\end{equation}
With a set of target hypotheses $\{h_i^T\}_{i=1}^{M}$, the optimization process can be expressed as jointly maximizing the expectation of MI given in Eq.~\ref{eq_MI} over the target hypotheses:
\begin{equation} \label{eq_target_loss}
\begin{split}
    & \max_{\psi^T} \mathbb{E}_{h \in \mathcal{H}^T} \left [ I(X^T; h(X^T)) \right ],
\end{split}
\end{equation}
where $\psi^T$ denotes the shared feature extractor among the $M$ target hypotheses and $h(X^T) := \hat{Y}^T$.
Similar to~\cite{liang2020we}, we fix the parameters of the classifiers for the target hypotheses (\ie, $f_i^T = f_i^S$) while updating $\psi^T$, due to the fact that both source and target domains share the same label space.

\subsection{Target hypothesis disparity regularization} \label{sec_hd}
In addition to the multiple hypotheses MI maximization (referred to as \textit{MI ensemble}) proposed in Eq.~\ref{eq_target_loss}, we introduce a hypothesis disparity regularization to better take advantage of the relationship among different hypotheses.
The proposed regularization is motivated by the crucial limitation of \textit{MI ensemble} where different hypotheses can be optimized in an unconstrained manner, resulting in undesirable disagreements on the target label predictions due to the unsupervised adaptation procedure. This is also in alignment with the finding of using MI for unsupervised discriminative clustering, where it is shown that a complexity penalty term is indispensable for MI maximization~\cite{krause2010discriminative}. The proposed regularization aims to take into account the uncertainty manifested through different hypotheses so as to marginalize out the undesirable disagreements resulted from \textit{MI ensemble}.


Here, we define hypothesis disparity (HD) as a dissimilarity measure of the predicted label probability distributions among different hypotheses over the input space $\mathcal{X}$:
\begin{equation} \label{eq_hd}
\begin{split}
    & \text{HD}_{h_i, h_j \in \mathcal{H}, i \neq j}(h_i, h_j) = \int_{\mathcal{X}} d(h_i(x), h_j(x)) p(x) dx,
\end{split}
\end{equation}
where $d(\cdot)$ can be any divergence measure between the predicted label probability distributions from the two hypotheses, \eg, $-\sum_K h_i(x) \log h_j(x)$ if using cross entropy as the divergence measure with $K$ unique labels. We discuss the relationship between cross entropy and Kullback–Leibler divergence as $d(\cdot)$ in MI maximization in the supplementary material, and provide empirical comparison in Table~\ref{tab:ablation}. We use cross entropy for $d(\cdot)$ throughout this paper.

We show with the empirical evidence that minimizing the HD among target hypotheses can effectively regularize the target hypotheses to maximally agree with each other, and help to coordinately learn better representation through the shared feature extractor resulting in better performance.

\subsection{HDMI} \label{sec_hdmi}
We now present our proposed approach---HDMI---by integrating the HD regularization into the \textit{MI ensemble}.
We denote $R(\mathcal{H})$ as a general form of regularization imposed on the hypothesis space $\mathcal{H}=\mathcal{H}^S \cup \mathcal{H}^T$.
Then we have the following objective function for MI-based unsupervised HTL:
\begin{equation} \label{eq_general_form}
\begin{split}
    & \mathbb{E}_{h \in \mathcal{H}^T} \left [-I(X^T; h(X^T)) \right ] + \lambda R(\mathcal{H}),
\end{split}
\end{equation}
and the proposed HDMI can be given as:
\begin{equation} \label{eq_hdmi}
\begin{split}
    & \mathbb{E}_{h \in \mathcal{H}^T} \left [-I(X^T; h(X^T)) \right ] + \lambda \mathbb{E}_{h_{i, j} \in \mathcal{H}^T, i \neq j} \left [\text{HD}(h_i, h_j) \right ].
\end{split}
\end{equation}
For computational efficiency, we set an anchor hypothesis that is randomly chosen from target hypotheses, and compute the average of the disparity between the anchor hypothesis and the rest $M-1$ hypotheses. 
In addition, if the HD among target hypotheses is minimized, the posterior predictive distribution in Eq.~\ref{eq_predictive} that computes the expectation of label predictions over the target hypothesis space can be equivalently simplified to the prediction from any hypothesis sampled from the target posterior:
\begin{equation} \label{eq_predictive_simplified}
    p(Y_t^*|\mathcal{D}_t, \mathcal{D}_s) \simeq p(Y_t^*|\mathcal{D}_t, h_t), h_t \sim p(h_t|\mathcal{D}_t, \{h_i^S\}_{i=1}^{M}).
\end{equation}
Therefore, we report the performance of the anchor hypothesis by using Eq.~\ref{eq_predictive_simplified} as our final HDMI performance, as compared with \textit{HDMI ensemble} that uses Eq.~\ref{eq_predictive} for the target predictive performance. Experimental results (Table~\ref{tab:office31} and Figure~\ref{fig:curve}) also empirically confirm the two are equivalent.

Our proposed HDMI with HD regularization is related to previous methods that use other forms of regularization: if $R(\mathcal{H}) = \bm{w}_t^\intercal\bm{w}_t$, with $\bm{w}_t$ denoting the parameters of a target hypothesis, we reach the regularized information maximization approach proposed for discriminative clustering in a single domain~\cite{krause2010discriminative}; if $R(\mathcal{H}) = \left \| \bm{w}_t - \bm{w}_s \right \|_2^2$, we obtain the typical $L_2$ regularization between the source and target hypotheses~\cite{kuzborskij2013stability,fernandes2019hypothesis}. We empirically find that the proposed HD regularization is superior to both of them (denoted as $L_2$ and $L_2 \textit{ source}$, respectively, in Table~\ref{tab:office31} and Table~\ref{tab:ablation}).



\section{Experiments}
\subsection{Setup}
We validate HDMI on three benchmark datasets for UDA in the context of HTL, and compare the adaptation/transfer performance with various state-of-the-art UDA and HTL methods as baselines.

\paragraph{Datesets} \textbf{Office-31}~\cite{saenko2010adapting} has three domains: Amazon (A), DSLR (D) and Webcam (W), with 31 classes and 4,652 images. \textbf{Office-Home}~\cite{venkateswara2017deep} is a more challenging dataset with 65 classes and 15,500 images in four domains: Artistic images (Ar), Clip art (Cl), Product images (Pr) and Real-World images (Rw). \textbf{VisDA-C}~\cite{peng2018visda} is a large-scale dataset with 12 classes, with 152,397 Synthetic images in the source domain and 55,388 Real images in the target domain.

\begin{table*}[!t]
    \centering
    \caption{Target accuracy ($\%$) on Office-31 with ResNet-50.}
    \begin{adjustbox}{width=0.90\textwidth}
      \begin{tabular}{c*{9}{c}}
        \toprule
        Source & \# of Hypotheses & Method & A$\rightarrow$D & A$\rightarrow$W & D$\rightarrow$A & D$\rightarrow$W & W$\rightarrow$A & W$\rightarrow$D & \fcolorbox{gray}{lightgray}{Avg.} \\
        \midrule
        \multirow{6}{*}{\cmark}
        & \multirow{6}{*}{single} & DAN~\cite{long2015learning} & 78.6 & 80.5 & 63.6 & 97.1 & 62.8 & 99.6 & \fcolorbox{gray}{lightgray}{80.4} \\
        & & DANN~\cite{ganin2015unsupervised} & 79.7 & 82.0 & 68.2 & 96.9 & 67.4 & 99.1 & \fcolorbox{gray}{lightgray}{82.2} \\
        & & SAFN+ENT~\cite{xu2019larger} & 90.7 & 90.1 & 73.0 & 98.6 & 70.2 & 99.8 & \fcolorbox{gray}{lightgray}{87.1} \\
        & & rRevGrad+CAT~\cite{deng2019cluster} & 90.8 & 94.4 & 72.2 & 98.0 & 70.2 & \textbf{100.} & \fcolorbox{gray}{lightgray}{87.6} \\
        & & CDAN+BSP~\cite{chen2019transferability} & 93.0 & 93.3 & 73.6 & 98.2 & 72.6 & \textbf{100.} & \fcolorbox{gray}{lightgray}{88.5} \\
        & & MDD~\cite{pmlr-v97-zhang19i} & 93.5 & \textbf{94.5} & \textbf{74.6} & 98.4 & 72.2 & \textbf{100.} & \fcolorbox{gray}{lightgray}{88.9} \\
        
        \midrule
        \multirow{8}{*}{\xmark}
        & \multirow{3}{*}{single} & Source only  &  79.7 & 75.7 & 61.2 & 96.0 & 59.8 & 98.2 & \fcolorbox{gray}{lightgray}{78.4} \\
        & & MI maximization  & 90.2 & 92.3 & 73.0 & 96.5 & 73.1 & 95.0 & \fcolorbox{gray}{lightgray}{86.7} \\
        
        & & SHOT~\cite{liang2020we} & 93.1 & 90.9 & 74.5 & 98.8 & 74.8 & 99.9 & \fcolorbox{gray}{lightgray}{88.7} \\
        
        \cmidrule(lr){2-10}
        & \multirow{6}{*}{multiple*} & Source only  & 81.1 & 77.2 & 61.2 & 96.5 & 60.7 & 98.4 & \fcolorbox{gray}{lightgray}{79.2}  \\
        & & MI ensemble  & 91.0 & 93.0 & 72.3 & 96.5 & 73.7 & 97.4 & \fcolorbox{gray}{lightgray}{87.3} \\
        
        & & MI ensemble + $L_2$ & 93.6 & 93.2 & 70.4 & 96.0 & 72.5 & 97.6 & \fcolorbox{gray}{lightgray}{87.2} \\
        & & MI ensemble + $L_2$ source & 92.0 & 91.7 & 68.7 & 97.9 & 66.1 & 99.8 & \fcolorbox{gray}{lightgray}{86.0} \\
        \cmidrule(lr){3-10}
        & & HDMI ($\lambda$=0.5) & \textbf{94.4} & 94.0 & 73.7 & \textbf{98.9} & \textbf{75.9} & 99.8 & \fcolorbox{gray}{lightgray}{\textbf{89.5}} \\
        & & HDMI ensemble ($\lambda$=0.5) & \textbf{94.4} & 94.0 & 73.6 & \textbf{98.9} & \textbf{75.9} & 99.8 & \fcolorbox{gray}{lightgray}{89.4} \\
        \bottomrule
        \multicolumn{5}{l}{\footnotesize * Two hypotheses as an illustration. More examples are shown in Table~\ref{tab:robustness}.}
      \end{tabular}
    \end{adjustbox}
    \label{tab:office31}
\end{table*}
\addtolength{\tabcolsep}{-4pt}
\begin{table*}[!t]
    \centering
    \caption{Target accuracy ($\%$) on Office-Home with ResNet-50.}
    \begin{adjustbox}{width=0.95\textwidth}
      \begin{tabular}{c*{13}{c}}
        \toprule
        Method & Ar$\rightarrow$Cl & Ar$\rightarrow$Pr & Ar$\rightarrow$Rw & Cl$\rightarrow$Ar & Cl$\rightarrow$Pr & Cl$\rightarrow$Rw & Pr$\rightarrow$Ar & Pr$\rightarrow$Cl & Pr$\rightarrow$Rw & Rw$\rightarrow$Ar & Rw$\rightarrow$Cl & Rw$\rightarrow$Pr & \fcolorbox{gray}{lightgray}{Avg.} \\
        \midrule
        DAN \cite{long2015learning} & 43.6 & 57.0 & 67.9 & 45.8 & 56.5 & 60.4 & 44.0 & 43.6 & 67.7 & 63.1 & 51.5 & 74.3 & \fcolorbox{gray}{lightgray}{56.3} \\
        DANN \cite{ganin2015unsupervised} & 45.6 & 59.3 & 70.1 & 47.0 & 58.5 & 60.9 & 46.1 & 43.7 & 68.5 & 63.2 & 51.8 & 76.8 & \fcolorbox{gray}{lightgray}{57.6} \\
		SAFN \cite{xu2019larger} & 52.0 & 71.7 & 76.3 & 64.2 & 69.9 & 71.9 & 63.7 & 51.4 & 77.1 & 70.9 & 57.1 & 81.5 & \fcolorbox{gray}{lightgray}{67.3} \\
		CDAN+TransNorm~\cite{wang2019transferable} & 50.2 & 71.4 & 77.4 & 59.3 & 72.7 & 73.1 & 61.0 & 53.1 & 79.5 & 71.9 & 59.0 & 82.9 & \fcolorbox{gray}{lightgray}{67.6} \\
        MDD~\cite{pmlr-v97-zhang19i} & 54.9 & 73.7 & 77.8 & 60.0 & 71.4 & 71.8 & 61.2 & 53.6 & 78.1 & 72.5 & 60.2 & 82.3 & \fcolorbox{gray}{lightgray}{68.1} \\
        \midrule
        
        SHOT-IM~\cite{liang2020we} & 52.8 & 72.9 & 78.4 & 65.4 & 73.8 & 74.1 & 64.6 & 50.8 & 78.9 & 72.7 & 53.5 & 81.2& \fcolorbox{gray}{lightgray}{68.3} \\
		SHOT~\cite{liang2020we} & 56.9 & \textbf{78.1} & 81.0 & \textbf{67.9} & 78.4 & 78.1 & \textbf{67.0} & 54.6 & 81.8 & 73.4 & 58.1 & \textbf{84.5} & \fcolorbox{gray}{lightgray}{71.6} \\
		
        \cmidrule(lr){2-14}
        Source only* & 45.6 & 69.2 & 76.5 & 55.3 & 64.4 & 67.4 & 55.1 & 41.6 & 74.4 & 66.0 & 46.3 & 79.4 & \fcolorbox{gray}{lightgray}{61.8}\\
        MI ensemble* & 55.2 & 71.9 & 80.2 & 62.6 & 76.8 & 77.8 & 63.2 & 53.8 & 81.1 & 67.9 & 58.3 & 81.4 & \fcolorbox{gray}{lightgray}{69.2} \\
        
        
        HDMI ($\lambda$=0.3)* & 57.4 & 76.9 & 81.6 & 67.6 & \textbf{79.1} & 78.1 & 65.1 & \textbf{56.0} & \textbf{82.5} & 73.5 & 59.5 & 83.6 & \fcolorbox{gray}{lightgray}{71.7}\\
                
        HDMI ($\lambda$=0.4)* & \textbf{57.8} & 76.7 & \textbf{81.9} & 67.1 & 78.8 & \textbf{78.8} & 66.6 & 55.5 & 82.4 & \textbf{73.6} & \textbf{59.7} & 84.0 & \fcolorbox{gray}{lightgray}{\textbf{71.9}} \\
        \bottomrule
        \multicolumn{13}{l}{\footnotesize * Two hypotheses as an illustration.}
      \end{tabular}
    \end{adjustbox}
    \label{tab:officehome}
    \vspace{-10pt}
\end{table*}
\addtolength{\tabcolsep}{4pt}

\begin{table}[!t]
    \centering
    \caption{Target domain per-class average accuracy~(\%) on VisDA-C (Synthetic$\rightarrow$Real) with ResNet-101.}
\scalebox{0.90}{
    \addtolength{\tabcolsep}{-8pt}
    \begin{tabular}[!t]{ccc}
        \toprule
        Source & Method & Avg. per-class accuracy \\ 
        \midrule
        \multirow{5}{*}{\cmark}
        & JAN~\cite{long2017deep} & 61.6 \\
        & GTA\cite{sankaranarayanan2018generate} & 69.5 \\
        & MCD~\cite{saito2018maximum} & 71.9 \\
        & CDAN~\cite{long2018conditional} & 70.0 \\
        & MDD~\cite{pmlr-v97-zhang19i} & 74.6 \\
        \midrule
        \multirow{5}{*}{\xmark}
        & SHOT-IM~\cite{liang2020we} & 77.9 \\
        & SHOT~\cite{liang2020we} & 79.6 \\ 
        
        \cmidrule(lr){2-3}
        & Source only & 44.6 \\
        & MI ensemble (two hypotheses) & 72.4  \\
        & HDMI (two hypotheses, $\lambda$=0.5) & \textbf{82.4} \\
        \bottomrule
    \end{tabular}
    \label{tab:visda_short}
    \addtolength{\tabcolsep}{8pt}
}
\end{table}

\paragraph{Baselines} The baseline methods can be divided into two categories depending on whether the model has access to both source and target domain data during adaptation. Most of the previous unsupervised domain adaptation methods (\eg, DAN~\cite{long2015learning}, DANN~\cite{ganin2015unsupervised}, rRevGrad+CAT~\cite{deng2019cluster}, CDAN+BSP~\cite{chen2019transferability}, CDAN+TransNorm \cite{wang2019transferable}, SAFN+ENT~\cite{xu2019larger}, MDD~\cite{pmlr-v97-zhang19i}) require \textit{source data access} during adaptation, whereas SHOT-IM~\cite{liang2020we} and SHOT~\cite{liang2020we} are unsupervised HTL methods without the \textit{source data access} constraint. We also report \textit{Source only}, which directly applies the source hypothesis to obtain the target predictions without any adaptation, and \textit{MI ensemble}, which uses multiple hypotheses for MI maximization but without the HD regularization.
In addition, we also report the results of two other regularization approaches, namely \textit{MI ensemble + $L_2$} and \textit{MI ensemble + $L_2$ source}. Our HDMI with independent classifiers (IC) from different random initialization is referred to as \textit{HDMI-IC}, so as to be distinguished from that with MC-dropout sampled classifiers denoted as \textit{HDMI-MC}.

\paragraph{Implementation details} We provide the details in the supplementary material. 
Note that we set the number of hypotheses $M=2$ by default unless otherwise stated, since empirical results suggest that the HD regularization between two hypotheses suffices HDMI for better performance.

\begin{table*}[!t]
    \centering
    \caption{Robustness of HDMI (Target accuracy ($\%$) on A$\rightarrow$D, Office-31). HDMI is robust to the choices of the number of hypotheses used ($M$) and weight hyperparameter tuning ($\lambda$).}
    \begin{adjustbox}{width=0.95\textwidth}
      \begin{tabular}{c*{9}{c}}
        \toprule
        \multirow{3}{*}{\# of Hypotheses ($M$)} & \multirow{3}{*}{Source only} & \multirow{3}{*}{MI maximization} & \multicolumn{6}{c}{\quad\quad\quad\quad\quad\quad HDMI} \\
        \cmidrule(lr){4-9}
        & & & \multicolumn{5}{c}{HDMI-IC} & HDMI-MC \\
        \cmidrule(lr){4-8} \cmidrule(lr){9-10}
        & & & $\lambda=0.1$ & $\lambda=0.3$ & $\lambda=0.5$ & $\lambda=0.7$ & $\lambda=1.0$ & $\lambda=0.5$ \\
        \midrule
        2 & 81.1 & 91.0 & 92.2 & 94.2 & 94.4 & 93.6 & 94.2 & 93.6 \\
        3 & 81.5 & 91.6 & 93.2 & 94.0 & 95.2 & 95.6 & 95.6 & 92.2 \\
        4 & 82.3 & 92.8 & 93.6 & 94.4 & 95.0 & 96.4 & 94.6 & 91.0 \\
        \bottomrule
      \end{tabular}
    \end{adjustbox}
    \label{tab:robustness}
\end{table*}

\begin{figure*}[!t]
    \centering
	\includegraphics[width=0.95\textwidth]{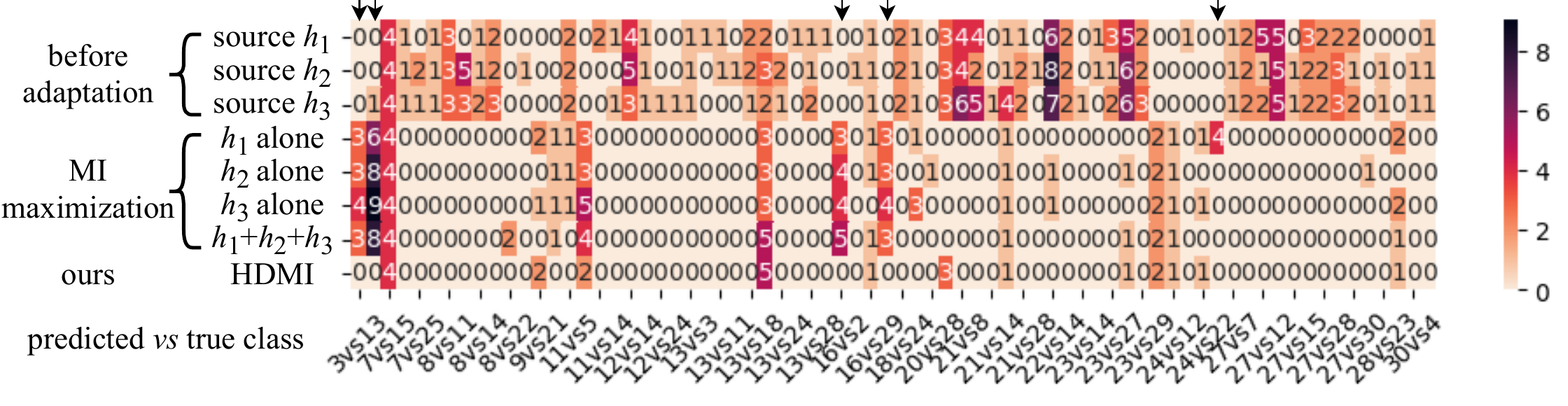}
	\vspace{-10pt}
	\caption{Target error analysis shows that HDMI (with three hypotheses) preserves more transferable source knowledge, as compared with using MI maximization alone (on A$\rightarrow$D, Office-31).}
	\label{fig:error}
\end{figure*}

\begin{figure}[!t]
    \centering
	\includegraphics[width=0.80\columnwidth]{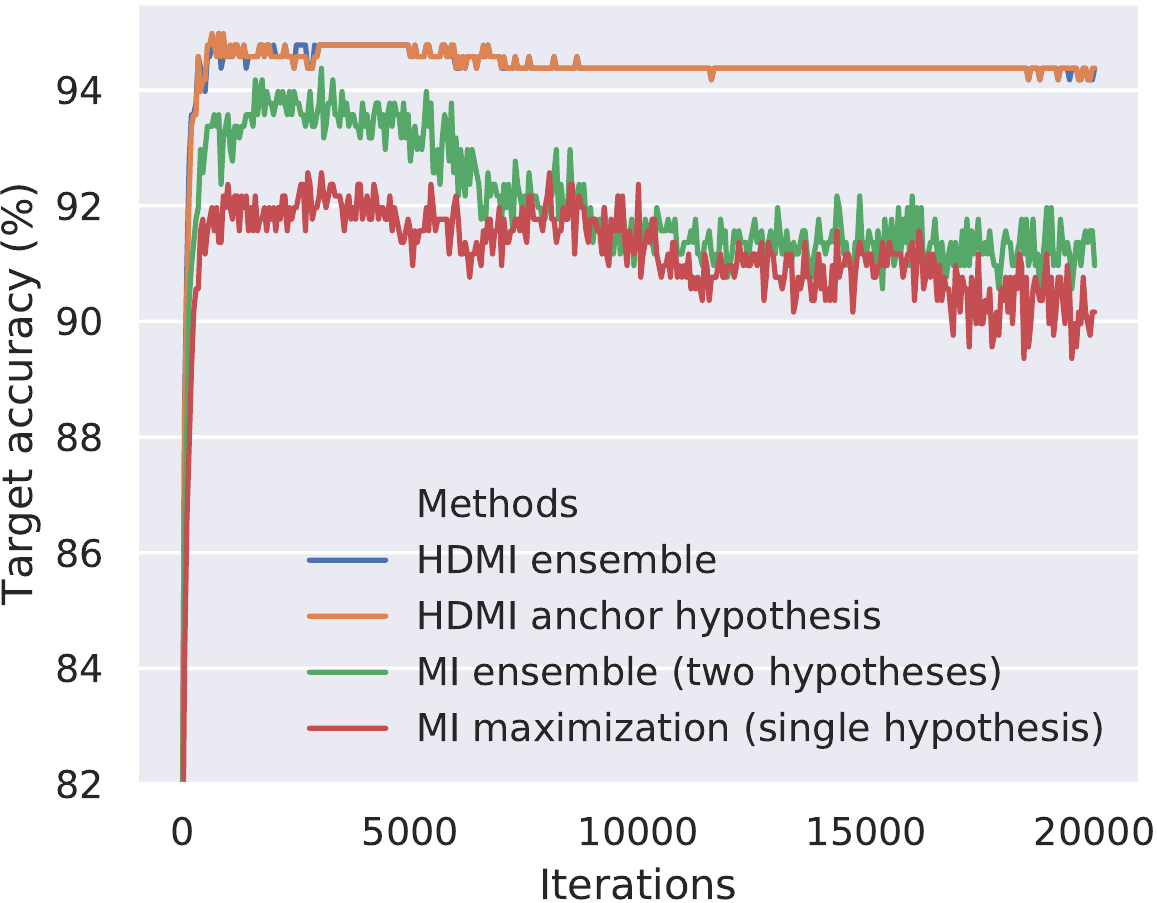}
	\caption{The hypothesis disparity regularization stabilizes the optimization for MI maximization (on A$\rightarrow$D, Office-31).}
	\vspace{-10pt}
	\label{fig:curve}
\end{figure}

\subsection{Results}
\paragraph{State-of-the-art performance of HDMI} We present the results of different methods on Office-31 in Table~\ref{tab:office31}, Office-Home in Table~\ref{tab:officehome}, and VisDA-C in Table~\ref{tab:visda_short}. The per-class accuracy for VisDA-C is detailed in Table~\ref{tab:visda_full} (supplementary material).
As seen from all tables, the proposed HDMI achieves state-of-the-art performance on the target domains in all datasets, even outperforming the methods that have additional access to the source data during adaptation (methods for which ``source'' marked as~\cmark~in the tables). In unsupervised HTL setting (methods for which ``source'' marked as~\xmark~in the tables), HDMI also outperforms previous state-of-the-art methods SHOT-IM (also based on MI maximization) and SHOT (with an extra pseudo-label based self-training strategy)~\cite{liang2020we}. Compared with \textit{MI ensemble}, adding the HD regularization effectively increases the target accuracy from 87.3$\%$ to 89.5$\%$ on Office-31, from 69.2$\%$ to 71.9$\%$ on Office-Home, and from 72.4$\%$ to 82.4$\%$ on VisDA-C. In addition, we also show in Table~\ref{tab:office31} that the proposed HD regularization in HDMI is superior to other forms of regularization such as those presented in \textit{MI ensemble + $L_2$} and \textit{MI ensemble + $L_2$ source}.

\paragraph{Robust performance of HDMI} To validate the robustness of HDMI in terms of the number of hypotheses $M$ and the hyperparameter $\lambda$, we perform experiments on A$\rightarrow$D, Office-31 with different configurations of $M$ and $\lambda$, and summarize the results in Table~\ref{tab:robustness}. It is shown that HDMI consistently obtains improved performance over the MI maximization baseline without the HD regularization. More importantly, we show that using two hypotheses suffices HDMI for the improved performance. We also find the implementation of HDMI with independent classifiers (HDMI-IC) is preferable to that with MC-droupout (HDMI-MC) due to its ability to cover different modes in the hypothesis space.

\subsection{Analyses}
Here, we investigate how multiple hypotheses and HD regularization improves the MI maximization process.

\begin{figure*}
    \centering
    \includegraphics[width=\textwidth]{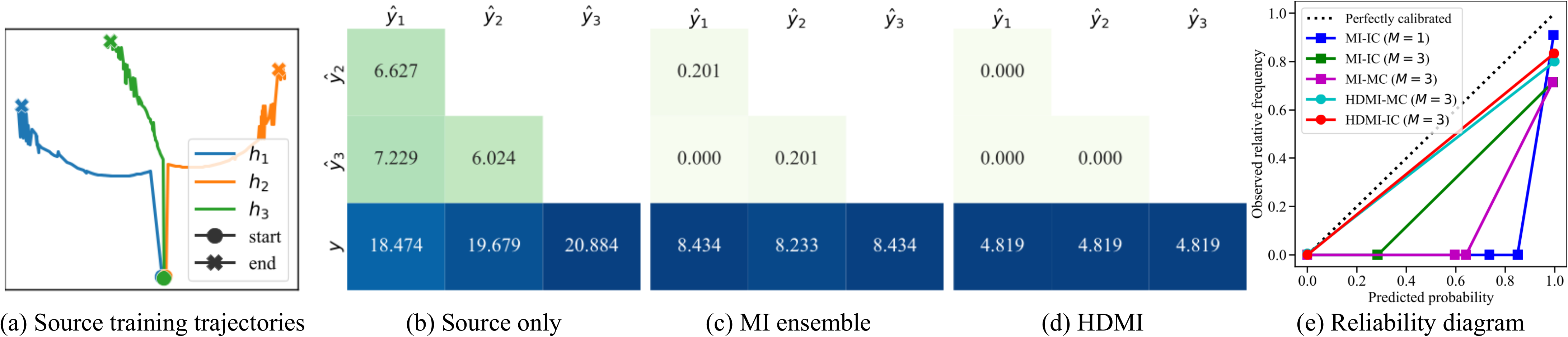}
    \caption{(a) t-SNE visualization comparing the trajectory of target predictions from different source hypotheses. We follow the same plotting procedure as in~\cite{fort2019deep}. (b-d) Disagreements between predictions from different hypotheses (\%), where $\hat{y}_i$ denotes the predictions of $h_i$ and $y$ denotes the ground-truth labels. (e) Reliability diagram of the target domain after transfer (class 11 as the positive class). All figures are on A$\rightarrow$D, Office-31.}
    \label{fig:disagreements}
    \vspace{-10pt}
\end{figure*}

\paragraph{The hypothesis disparity regularizes MI maximization} Figure~\ref{fig:curve} shows the learning curves of the target accuracy for different mutual information based approaches with or without the HD regularization.
As shown in the figure, the target performance degrades in ``MI maximization (single hypothesis)'' due to the lack of proper regularization.
Furthermore, we find that the use of deep ensemble in ``MI ensemble (two hypotheses)'' does not help alleviate this performance degradation problem.
This necessitates the proposal of our HD regularized MI maximization, where the transfer process is stable and effective.

\paragraph{HDMI facilitates the positive transfer of multiple modes from the source hypotheses}
Figure~\ref{fig:error} summarizes the fine-grained hypothesis-level prediction errors made by different approaches.
The figure reveals that direct MI maximization (row 4-7) suffers from negative transfer and introduces new errors that were not present in the \textit{Source only} models (row 1-3) before adaptation, \eg, columns with arrows, indicating partial lost of the transferable source knowledge during adaptation.
In contrast, HDMI (row 8) facilitates positive transfer of different modes learned from the source domain, shown in Figure~\ref{fig:disagreements}~(a), and results in stable and effective target adaptation.
As a result, the HD regularization prevents negative transfer from the MI maximization and facilitates the positive transfer of multiple modes from source hypotheses.

\paragraph{HDMI maximally reduces the disagreement among target hypotheses} Figure~\ref{fig:disagreements} (b)-(d) compare the disagreement among predictions from different target hypotheses and the ground-truth, where HDMI (Figure~\ref{fig:disagreements} (d)) is shown to maximally reduce the disagreement compared with \textit{Source only} (Figure~\ref{fig:disagreements} (b)) and \textit{MI ensemble} (Figure~\ref{fig:disagreements} (c)), demonstrating the effectiveness of the HD regularization in bringing target hypotheses to align with each other. We have similar findings on the KL divergence of example-level predictions between target hypotheses (Figure~\ref{fig:disagreement_kl}, supplementary material).

\paragraph{HDMI presents well-calibrated predictive uncertainty}
Uncertainty calibration is especially important for the performance of hypothesis transfer between different source and target domains where better calibrated probabilities lead to more effective hypothesis transfer.
To investigate whether HDMI benefits from uncertainty calibration, we plot the reliability diagram~\cite{degroot1983comparison,niculescu2005predicting} of different approaches and confirm that HDMI is better calibrated than other approaches (Figure~\ref{fig:disagreements} (e)). In consistent with~\cite{lakshminarayanan2017simple}, we also find that multiple hypotheses using independent classifiers~(IC) is superior to that with MC-dropout sampled classifiers (MC) in both cases of \textit{MI ensemble} and HDMI. Quantitative analysis of the uncertainty calibration confirms that HDMI has the best Brier score and the expected calibration error (ECE) score~\cite{naeini2015obtaining} (Table~\ref{tab:calibration}, supplementary material).



\begin{table}[!t]
    \centering
    \caption{Ablation study (on Office-31).}
    \scalebox{0.95}{
    \renewcommand{\arraystretch}{1.1}
    \addtolength{\tabcolsep}{-7pt}

    \begin{tabular}{c*{6}{c}}
        \toprule
        Method* & Target avg. accuracy ($\%$) \\
        \midrule
        Source only & 79.2 \\
        MI ensemble & 87.3 \\
        HDMI & \textbf{89.5} \\
        \midrule
        HDMI with KL & 88.6 \\
        \hdashline
        MI ensemble (independent $\psi$) & 87.7 \\
        HDMI (independent $\psi$) & 87.5 \\
        \hdashline
        HD only & 84.8 \\ 
        Conditional Entropy + HD & 85.7 \\
        \hdashline
        MI ensemble + $L_2$ & 87.2 \\
        MI ensemble + $L_2$ source & 86.0 &  
        \\

        \bottomrule
        \multicolumn{5}{l}{\footnotesize * With two hypotheses, $\lambda$=0.5.}
    \end{tabular}
    \label{tab:ablation}
    \addtolength{\tabcolsep}{7pt}
}
\end{table}

\subsection{Ablation study}
We summarize the results of ablation studies in Table~\ref{tab:ablation}.
We first evaluate the impact of shared feature extractor $\psi$ among target hypotheses by comparing ``MI ensemble (independent $\psi$)'' with ``HDMI (independent $\psi$)''. We find that the HD regularization does not help MI maximization if the feature extractors are independent, suggesting that HD regularization works through learning better representations shared by different target hypotheses.
In addition, we also find MI maximization performs better than conditional entropy minimization in unsupervised HTL, similar to the finding in discriminative clustering~\cite{krause2010discriminative}. Lastly, we show cross entropy measure surrogates the proposed hypothesis disparity better than KL divergence. The detailed results are provided in the supplementary material (Table~\ref{tab:ablation_sharedORindependent}, Table~\ref{tab:ablation_condEntropy} and Table~\ref{tab:ablation_CEorKL}).

\section{Conclusion}
In this paper, we tackle the problem of unsupervised hypothesis transfer learning to bridge the gap between unsupervised domain adaptation and hypothesis transfer learning. We propose a hypothesis disparity regularized mutual information maximization approach that not only employs multiple source and target hypotheses but also utilizes the relationship among different hypotheses to overcome the limitation of mutual information maximization with a single source and target hypothesis.
Empirical results demonstrate that the proposed hypothesis disparity regularization minimizes undesirable disagreements among hypotheses and preserves more transferable knowledge from the source domain.
Our approach achieves state-of-the-art performance on three benchmark datasets for unsupervised domain adaptation in the context of hypothesis transfer learning.

\bibliographystyle{aaai}
\bibliography{main}

\clearpage
\onecolumn
\section{Supplementary Material}
\subsection{Implementation details} \label{sec_implementation}
\normalsize
Similar to previous work~\cite{pmlr-v97-zhang19i,liang2020we}, we adopt ResNet~\cite{he2016deep} models (\eg, ResNet-50 for office datasets and ResNet-101 for the VisDA-C dataset) pretrained on ImageNet~\cite{russakovsky2015imagenet} as the backbone inside the feature extractor $\psi$. The ResNet backbone is then followed by a bottleneck layer (a FC layer with BN, ReLU and Dropout). The extracted feature dimension is 1024 for Office-31, and 2048 for both Office-Home and VisDA-C. We follow the same architecture choices as in~\cite{pmlr-v97-zhang19i} for the classifiers $\{f_i\}_{i=1}^{M}$, \ie, two-layer neural networks (two FC layers with ReLU and Dropout). 

To train the source hypotheses, we use learning rate 3e-4 with batch size 32 for 5k iterations. The target hypotheses are trained with a larger batch size 64, learning rate 3e-4 (Office-31), 1e-3 (Office-Home) and 1e-4 (VisDA-C) for 20k iterations (40k for VisDA-C). The hyperparameter $\lambda$ is set to 0.5 and the number of hypotheses $M$ is set to 2 by default. We use SGD optimizer with nesterov momentum 0.9 and weight decay 5e-4 for training both source and target hypotheses. Following~\cite{pmlr-v97-zhang19i,liang2020we}, the ResNet backbone in the feature extractor $\psi$ is fine-tuned with a 10 times smaller learning rate.


\subsection{Detailed per-class accuracy on VisDA-C (Table~\ref{tab:visda_full})}
\addtolength{\tabcolsep}{-3pt}
\begin{table*}[htbp]
	\centering
	\caption{Target accuracy (\%) on VisDA-C (Synthetic$\rightarrow$Real) with ResNet-101.}
	\begin{adjustbox}{width=\textwidth}
		\begin{tabular}{lccccccccccccc}
		\toprule
		Method & aeroplane & bicycle & bus & car & horse & knife & motorcycle & person & plant & skateboard & train & truck & \fcolorbox{gray}{lightgray}{Per-class} \\
		\midrule
		ResNet-101 \cite{he2016deep}  & 55.1 & 53.3 & 61.9 & 59.1 & 80.6 & 17.9 & 79.7 & 31.2  & 81.0 & 26.5  & 73.5 & 8.5  & \fcolorbox{gray}{lightgray}{52.4}   \\
		DANN \cite{ganin2015unsupervised}   & 81.9 & 77.7 & 82.8 & 44.3 & 81.2 & 29.5 & 65.1 & 28.6  & 51.9 & 54.6  & 82.8 & 7.8  & \fcolorbox{gray}{lightgray}{57.4}   \\
		DAN \cite{long2015learning}   & 87.1 & 63.0 & 76.5 & 42.0 & 90.3 & 42.9 & 85.9 & 53.1  & 49.7 & 36.3  & 85.8 & 20.7 & \fcolorbox{gray}{lightgray}{61.1}   \\
		ADR \cite{saito2017adversarial} & 94.2 & 48.5 & 84.0 & \textbf{72.9} & 90.1 & 74.2 & \textbf{92.6} & 72.5 & 80.8 & 61.8 & 82.2 & 28.8 & \fcolorbox{gray}{lightgray}{73.5} \\
		CDAN \cite{long2018conditional}   & 85.2 & 66.9 & 83.0 & 50.8 & 84.2 & 74.9 & 88.1 & 74.5  & 83.4 & 76.0  & 81.9 & 38.0 & \fcolorbox{gray}{lightgray}{73.9}   \\
		CDAN+BSP \cite{chen2019transferability} & 92.4 & 61.0 & 81.0 & 57.5 & 89.0 & 80.6 & 90.1 & 77.0 & 84.2 & 77.9 & 82.1 & 38.4 & \fcolorbox{gray}{lightgray}{75.9} \\
		SAFN \cite{xu2019larger} & 93.6 & 61.3 & \textbf{84.1} & 70.6 & \textbf{94.1} & 79.0 & 91.8 & \textbf{79.6} & \textbf{89.9} & 55.6 &  \textbf{89.0} & 24.4 & \fcolorbox{gray}{lightgray}{76.1} \\
		SWD \cite{lee2019sliced} & 90.8 & 82.5 & 81.7 & 70.5 & 91.7 & 69.5 & 86.3 & 77.5 & 87.4 & 63.6 & 85.6 & 29.2 & \fcolorbox{gray}{lightgray}{76.4} \\
		SHOT-IM~\cite{liang2020we} & 89.9 & 80.1 & 79.1 & 50.9 & 88.0 & 90.5 & 78.2 & 78.5 & 89.3 & 80.2 & 85.8 & 44.9 & \fcolorbox{gray}{lightgray}{77.9} \\
		SHOT~\cite{liang2020we} & 92.6 & 81.1 & 80.1 & 58.5 & 89.7 & 86.1 & 81.5 & 77.8 & 89.5 & 84.9 & 84.3 & 49.3 & \fcolorbox{gray}{lightgray}{79.6} \\
		\midrule

		Source only* & 58.3  & 12.1  & 45.2  & 60.9  & 64.7  & 9.1  & 88.5  & 12.1  & 62.0  & 30.9  & 86.7  & 4.7  & \fcolorbox{gray}{lightgray}{44.6} \\
		MI ensemble*   & \textbf{95.3}  & 84.3  & 76.2  & 47.5  & 91.0  & 22.5  & 73.6  & 75.6  & 82.0  & 76.2  & 87.6  & \textbf{56.7}  & \fcolorbox{gray}{lightgray}{72.4}  \\
		HDMI ($\lambda$=0.5)* & 94.7  & \textbf{85.1}  & 80.3  & 62.3  & 92.1  & \textbf{95.5}  & 86.4  & 77.8  & 84.8  & \textbf{85.7}  & 88.4  & 55.4  & \fcolorbox{gray}{lightgray}{\textbf{82.4}} \\
		\bottomrule
        \multicolumn{13}{l}{\footnotesize * With two hypotheses.}
		\end{tabular}
	\end{adjustbox}
	\label{tab:visda_full}
\end{table*}
\addtolength{\tabcolsep}{3pt}

\begin{figure}[htbp]
    \centering
    \includegraphics[width=0.95\textwidth]{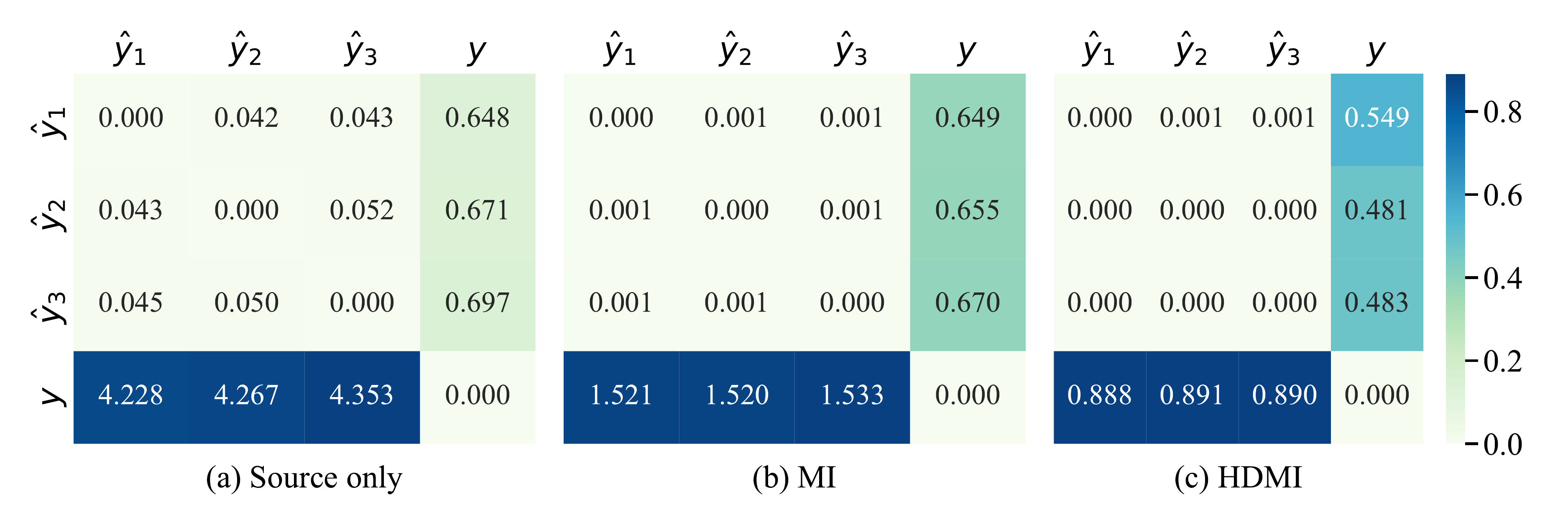}
    \caption{KL divergence between label probabilities from different hypotheses on A$\rightarrow$D, Office-31. The columns represent $p$ and the rows represent $q$ in $\mathrm{KL}[p\parallel q]$.}
    \label{fig:disagreement_kl}
\end{figure}

\subsection{Disagreement analysis of the label probabilities based on KL divergence (Figure~\ref{fig:disagreement_kl})}

\subsection{Quantitative analysis on uncertainty calibration (Table~\ref{tab:calibration})}
\begin{table}[!ht]
\caption{Evaluation of uncertainty calibration on the target domain (A$\rightarrow$D, Office-31).}
    \centering
    \begin{tabular}{ccccc}
    \toprule
    & Classifiers & Number of hypotheses & Brier score & ECE \\\midrule
    \multirow{3}{*}{ Source only } & independent & 1 & 0.2898 & 0.0124 \\
    & MC-dropout & 3 & 0.3171 & 0.0133 \\
    & independent & 3 & 0.2763 & 0.0116 \\\midrule
    \multirow{3}{*}{ MI } & independent & 1 & 0.1634 & 0.0057 \\
    & MC-dropout & 3 & 0.1584 & 0.0053 \\
    & independent & 3 & 0.1593 & 0.0054 \\\midrule
    \multirow{2}{*}{ HDMI } & MC-dropout & 3 & 0.1566 & 0.0051 \\
    & independent & 3 & \textbf{0.0961} & \textbf{0.0031} \\\bottomrule
    \end{tabular}
    
    \label{tab:calibration}
\end{table}

\subsection{Detailed ablation study on feature extractor (Table~\ref{tab:ablation_sharedORindependent})}
\begin{table*}[htbp]
\renewcommand{\arraystretch}{1.2}
    \centering
    \caption{Comparing shared feature extractor with independent feature extractor. Target accuracy ($\%$) on Office-31.}
    \begin{adjustbox}{width=0.9\textwidth}
      \begin{tabular}{c*{8}{c}}
        \toprule
        Method* & A$\rightarrow$D & A$\rightarrow$W & D$\rightarrow$A & D$\rightarrow$W & W$\rightarrow$A & W$\rightarrow$D & \fcolorbox{gray}{lightgray}{Avg.} \\
        \midrule
        MI ensemble (independent feature extractor)   & 92.4 & 92.7 & 70.7 & 98.7 & 71.9 & \textbf{100.} & \fcolorbox{gray}{lightgray}{87.7} \\
        MI ensemble (shared feature extractor)  & 91.0 & 93.0 & 72.3 & 96.5 & 73.7 & 97.4 & \fcolorbox{gray}{lightgray}{87.3} \\\midrule
        HDMI (independent feature extractor) & 91.5 & 92.3 & 71.9 & 98.6 & 71.1  & 99.6 & \fcolorbox{gray}{lightgray}{87.5} \\
        HDMI (shared feature extractor) & \textbf{94.4} & \textbf{94.0} & \textbf{73.7} & \textbf{98.9} & \textbf{75.9} & 99.8 & \fcolorbox{gray}{lightgray}{\textbf{89.5}} \\
        \bottomrule
        \multicolumn{8}{l}{\footnotesize * with two hypotheses, and $\lambda=0.5$ for HD.}
      \end{tabular}
    \end{adjustbox}
    \label{tab:ablation_sharedORindependent}
\end{table*}

\subsection{Detailed ablation study on conditional entropy minimization (Table~\ref{tab:ablation_condEntropy} and Table~\ref{tab:ablation_condEntropy_finetune})}
\begin{table*}[htbp]
    \centering
    \caption{Comparing MI maximization and conditional entropy minimization, both with HD regularization. Target accuracy ($\%$) on Office-31.}
    \begin{adjustbox}{width=0.9\textwidth}
      \begin{tabular}{c*{8}{c}}
        \toprule
        Method & A$\rightarrow$D & A$\rightarrow$W & D$\rightarrow$A & D$\rightarrow$W & W$\rightarrow$A & W$\rightarrow$D & \fcolorbox{gray}{lightgray}{Avg.} \\
        \midrule
        Source only (single hypothesis) &  79.7 & 75.7 & 61.2 & 96.0 & 59.8 & 98.2 & \fcolorbox{gray}{lightgray}{78.4} \\
        MI maximization (single hypothesis) & 90.2 & 92.3 & 73.0 & 96.5 & 73.1 & 95.0 & \fcolorbox{gray}{lightgray}{86.7} \\
        Conditional Entropy minimization (single hypothesis) & 91.0 & 91.6 & 61.8 & 98.5 & 60.1 & \textbf{100.} & \fcolorbox{gray}{lightgray}{83.8} \\
        \midrule
        
        HD only* & 93.4 & 90.4 & 65.7 & 98.5 & 60.8 & 99.8 & \fcolorbox{gray}{lightgray}{84.8} \\ 

        Conditional Entropy ensemble* & \textbf{96.6} & 91.1 & 67.3 & 98.5 & 62.1 & 99.8 & \fcolorbox{gray}{lightgray}{85.9} \\
        Conditional Entropy + HD* & 95.0 & 90.8 & 68.8 & 98.5 & 61.2 & 99.8 & \fcolorbox{gray}{lightgray}{85.7} \\
        MI ensemble* & 91.0 & 93.0 & 72.3 & 96.5 & 73.7 & 97.4 & \fcolorbox{gray}{lightgray}{87.3} \\
        
        HDMI* & 94.4 & \textbf{94.0} & \textbf{73.7} & \textbf{98.9} & \textbf{75.9} & 99.8 & \fcolorbox{gray}{lightgray}{\textbf{89.5}} \\
        \bottomrule
        \multicolumn{8}{l}{\footnotesize * with two hypotheses, and $\lambda=0.5$ for HD. Fine-tuning on HD regularized conditional entropy minimization still does not} \\
        \multicolumn{8}{l}{\footnotesize outperform HDMI (Table~\ref{tab:ablation_condEntropy_finetune}, supplementary material)}
      \end{tabular}
    \end{adjustbox}
    \label{tab:ablation_condEntropy}
\end{table*}

\begin{table}[ht!]
    \centering
    \caption{Fine-tuning on hypothesis disparity regularized conditional entropy minimization does not outperform HDMI. }
    \scalebox{0.95}{
      \begin{tabular}{c*{9}{c}}
        \toprule
        HDMI & \multicolumn{5}{c}{Conditional Entropy + HD} \\
        \cmidrule(lr){1-1} \cmidrule(lr){2-8}
        $\lambda=0.5$ & $\lambda=0.1$ & $\lambda=0.3$ & $\lambda=0.5$ & $\lambda=0.7$ & $\lambda=1.0$\\
        \midrule
        \textbf{94.0} & 90.8 & 90.9 & 90.8 & 90.9 & 88.8 \\
        \bottomrule
      \end{tabular}
    }
    \label{tab:ablation_condEntropy_finetune}
\end{table}


\subsection{HDMI with cross entropy or Kullback–Leibler divergence for HD} \label{sec_suppl_CEorKL}
\normalsize
Here, we present the relationship between two different implementations of HDMI, \ie, with either cross entropy (CE) or Kullback–Leibler (KL) divergence for the HD regularization. As shown in the following equation (Eq.~\ref{eq_hdmi_ce_kl}), the objective function for the target training using CE ($L_{target}^{\text{CE based}}$) can be viewed as that of using KL divergence ($L_{target}^{\text{KL based}}$) with additional emphasis on the predictive confidence of the \textit{anchor} hypothesis $h_i$, where $i \in [1, M]$:
\begin{equation} \label{eq_hdmi_ce_kl}
\begin{aligned}
    L_{target}^{\text{CE based}} & = \sum_{m=1}^M -I(X^T; \hat{Y}^T_m) + \lambda \sum_{j \neq i} \mathbb{E}_{x \in \mathcal{X}^T} \left [ H(h_i(x), h_j(x)) \right ] \\
    & = \sum_{m=1}^M -I(X^T; \hat{Y}^T_m) + \lambda \sum_{j \neq i} \mathbb{E}_{x \in \mathcal{X}^T} \left [ - \sum_{K} h_i(x) \log h_j(x) \right ] \\
    & = \sum_{m=1}^M -I(X^T; \hat{Y}^T_m) + \lambda \sum_{j \neq i} \mathbb{E}_{x \in \mathcal{X}^T} \left [ - \sum_{K} h_i(x) \log h_i(x) - \sum_{K} h_i(x) \log \left( \frac{h_j(x)}{h_i(x)} \right) \right ] \\
    & = \sum_{m=1}^M \left ( H(\hat{Y}^T_m | X^T) - H(\hat{Y}^T_m) \right ) + \lambda  \sum_{j \neq i} \left ( H(\hat{Y}^T_i|X^T) + \mathbb{E}_{x \in \mathcal{X}^T} \left [ D_{KL}(h_i(x) \vert \vert h_j(x)) \right] \right ) \\
    & = \sum_{m=1}^M \left ( H(\hat{Y}^T_m | X^T) - H(\hat{Y}^T_m) \right ) + \lambda (M-1) H(\hat{Y}^T_i|X^T) + \lambda  \sum_{j \neq i} \mathbb{E}_{x \in \mathcal{X}^T} \left [ D_{KL}(h_i(x) \vert \vert h_j(x)) \right ] \\
    & = L_{target}^{\text{KL based}} + \lambda (M-1) H(\hat{Y}^T_i|X^T).
\end{aligned}
\end{equation}

The empirical comparisons are given in Table~\ref{tab:ablation_CEorKL}.
\begin{table*}[htbp]
    \centering
    \caption{Comparing cross entropy and KL divergence for HD. Target accuracy ($\%$) on Office-31.}
      \begin{tabular}{c*{8}{c}}
        \toprule
        Method* & A$\rightarrow$D & A$\rightarrow$W & D$\rightarrow$A & D$\rightarrow$W & W$\rightarrow$A & W$\rightarrow$D & \fcolorbox{gray}{lightgray}{Avg.} \\
        \midrule
        
        HDMI with KL & 93.2 & 93.5 & \textbf{73.8} & 97.0 & 74.3 & \textbf{99.8} & \fcolorbox{gray}{lightgray}{88.6} \\
        HDMI with CE  & \textbf{94.4} & \textbf{94.0} & 73.7 & \textbf{98.9} & \textbf{75.9} & \textbf{99.8} & \fcolorbox{gray}{lightgray}{\textbf{89.5}} \\
        \bottomrule
        \multicolumn{8}{l}{\footnotesize * with two hypotheses, and $\lambda=0.5$ for HD.}
      \end{tabular}
    \label{tab:ablation_CEorKL}
\end{table*}




\end{document}